\documentclass[letterpaper, 10 pt, conference]{ieeeconf}
\IEEEoverridecommandlockouts
\let\labelindent\relax
\usepackage{enumitem}
\usepackage{balance}
\usepackage[font={small}]{caption}
\usepackage{subcaption}
\usepackage{array}
\usepackage{textcomp}
\usepackage{mathtools, nccmath}
\usepackage{graphicx}
\usepackage{amsfonts}
\usepackage{amsmath}
\usepackage{amssymb}
\usepackage{algorithm}
\usepackage{algorithmic}
\usepackage[absolute,overlay]{textpos}
\usepackage{hyperref}
\usepackage{tikz}
\usepackage{arydshln}
\usepackage{multirow}
\usepackage{bm}
\usepackage{epstopdf}
\usepackage{cite}
\usepackage{siunitx}
\usepackage{multirow}
\usepackage{lipsum}

\usepackage[dvipsnames]{xcolor}
\usepackage{colortbl}
\definecolor{light-gray}{gray}{0.93}
\definecolor{semilight-gray}{gray}{0.8}

\usepackage{amsthm}

\newtheorem{theorem}{Theorem}

\newtheorem{remark}{Remark}
\theoremstyle{definition}
\newtheorem{definition}{Definition}
\newtheorem{problem}{Problem}

\newtheorem{proposition}{Proposition}

\title{\LARGE \bf
Iterative Convex Optimization with Control Barrier Functions for Obstacle Avoidance among Polytopes
}

\author{Shuo Liu$^{1*}$, Zhe Huang$^{1*}$ and Calin A. Belta$^{2}$% <-this % stops a space
\thanks{$^{*}$These authors contributed equally to this work.}%
\thanks{This work was supported in part by the NSF under grant IIS-2024606
at Boston University and by a Brendan Iribe endowed professorship at the
University of Maryland.}
\thanks{$^{1}$S. Liu and Z. Huang are with Boston
University, Brookline, MA, USA {\tt\small \{liushuo, huangz7\}@bu.edu}}
\thanks{$^{2}$C. Belta is with the Department of Electrical and Computer Engineering and with the Department of Computer Science, University of Maryland, College Park, MD, USA 
        {\tt\small cbelta@umd.edu}}%
\thanks{Implementation code is released on \url{https://github.com/jh01231230/impc-cbf-polytope-obstacle-avoidance}.}
}

\begin{document} 
\maketitle

\begin{abstract}
Obstacle avoidance of polytopic obstacles by polytopic robots  is a challenging problem in optimization-based control and trajectory planning.
Many existing methods rely on smooth geometric approximations, such as hyperspheres or ellipsoids, which allow differentiable distance expressions but distort the true geometry and restrict the feasible set. 
Other approaches integrate exact polytope distances into nonlinear model predictive control (MPC), resulting in nonconvex programs that limit real-time performance.
In this paper, we construct linear discrete-time control barrier function (DCBF) constraints by deriving supporting hyperplanes from exact closest-point computations between convex polytopes. 
We then propose a novel iterative convex MPC-DCBF framework, where local linearization of system dynamics and robot geometry ensures convexity of the finite-horizon optimization at each iteration. 
The resulting formulation reduces computational complexity and enables fast online implementation for safety-critical control and trajectory planning of general nonlinear dynamics. 
The framework extends to multi-robot and three-dimensional environments. 
Numerical experiments demonstrate collision-free navigation in cluttered maze scenarios with millisecond-level solve times.
\end{abstract}

\section{Introduction}
\label{sec:Introduction}
Safety-critical optimal control is a fundamental problem in robotics. Steering a system to a goal while avoiding obstacles and minimizing energy consumption can be formulated as a constrained optimal control problem using continuous-time Control Barrier Functions (CBFs) \cite{ames2019control, ames2016control}. By solving the resulting Quadratic Program (QP) at each sampling instant, 
the CBF-based safety constraints can be enforced in real time. Subsequent developments in CBF-based safety-critical control have extended the framework to handle high relative degree constraints \cite{nguyen2016exponential,xiao2021high, liu2023auxiliary}, mixed relative degree constraints \cite{liu2025auxiliary}, and  Discrete-time CBFs (DCBFs) \cite{agrawal2017discrete}.

CBFs have been widely used for obstacle avoidance across a range of applications \cite{ khazoom2022humanoid,liu2025safety,liu2025learning,chen2017obstacle,verginis2019closed}. However, many existing approaches approximate robots and/or obstacles using smooth geometric shapes such as hyperspheres \cite{chen2017obstacle} and ellipsoids \cite{verginis2019closed}. While these representations admit closed-form and differentiable distance expressions that facilitate real-time optimization, they may not accurately reflect the true geometry of complex obstacles. Alternatively, some methods construct separating constraints based on tangent hyperplanes of obstacle boundaries \cite{liu2025safety,liu2025learning}. For nonconvex obstacles, tangent hyperplanes at concave boundary regions can intersect the obstacle interior, potentially leading to unsafe characterizations of the feasible set. Polytopic representations more precisely capture geometric structure and admit separating hyperplanes that do not intersect the interior, ensuring safe separation.
\begin{figure}
\vspace{3mm}
    \centering
    \includegraphics[scale=0.12]{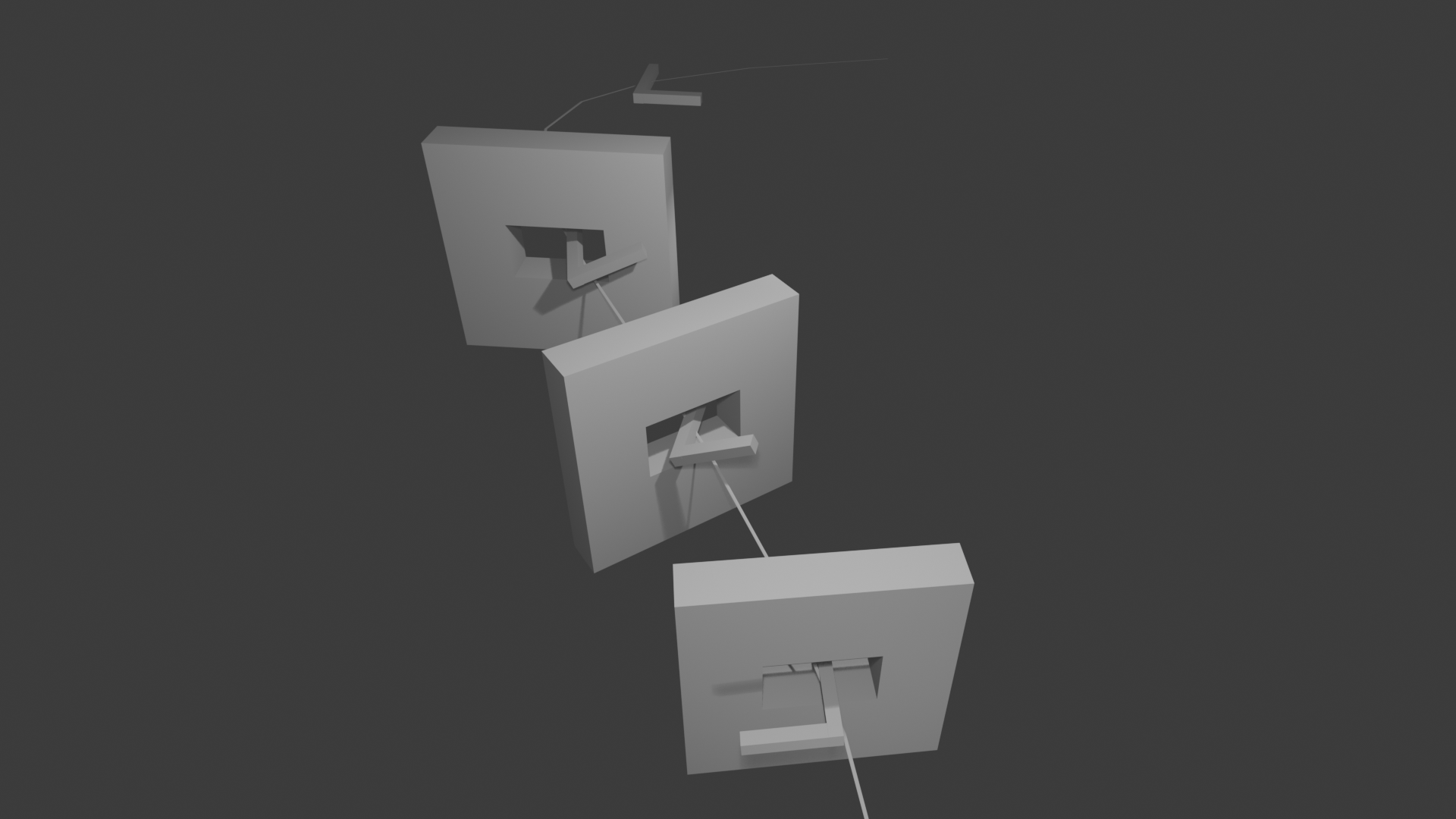}
    \caption{Motivating example: an L-shaped robot performing narrow-passage traversal. Simulation results are shown in Sec. \ref{sec:3D-sim}.}
    \label{fig:snapshot}
\end{figure}

CBF constructions for polytopes have been explored in several works. The method in \cite{singletary2022safety} employs the Signed Distance Function (SDF) 
with a local linear approximation of the CBF gradient; although approximation 
errors are compensated to ensure safety, the linearization introduces conservatism, 
especially near nonsmooth boundaries. 
An optimization-free smooth approximate-SDF CBF is proposed in \cite{wu2025optimization}, 
where conservatism is reduced via hyperparameter tuning, but geometric 
over-approximation remains inherent in the approximate distance representation. 
In \cite{wei2024diffocclusion,peng2023safe}, polygonal obstacles are replaced 
by smooth surrogate geometries—via differentiable scaling optimization or 
polynomial fitting—to enable gradient computation. While computationally 
convenient, such smooth approximations may distort the true geometry and 
restrict the feasible set.

\begin{table*}[t!]
    \centering
 \vspace*{3mm}   \caption{Conceptual comparison of existing CBF-based collision avoidance methods for polytopic sets. \textbf{Implicit} indicates that CBFs are defined via optimization. \textbf{Smoothness} refers to the differentiability of the CBF function. \textbf{Metric} describes the geometric quantity used to quantify separation in the CBF. 
    }
    \vspace{-0.5em}
    \resizebox{0.98\textwidth}{!}{
    \begin{tabular}{|c|ccccccc|}
    \hline
        %\rowcolor{semilight-gray} 
         \textbf{Approaches}
        &\textbf{Explicitness}
        &\textbf{Smoothness}
        & \textbf{Metric}
        & \textbf{Optimization} %Opti-defined CBF
        & \textbf{Horizon}
        & \textbf{3-D Demonstrated}
        & \textbf{Multi-robot Demonstrated}\\
    \hline 
        SDF-linearized CBF \cite{singletary2022safety}
        & Implicit 
        & Smooth
        & Distance
        & Convex
        & $1$
        & Yes
        & No\\
    \hline
        Opti-free CBF \cite{wu2025optimization}
        & Explicit 
        & Smooth
        & Distance
        & Convex
        & $1$
        & No
        & Yes\\
    \hline
        Diff-opti CBF \cite{wei2024diffocclusion}
        & Implicit 
        & Smooth
        & Scaling factor
        & Convex %Conservative
        & $1$
        & Yes 
        & No\\
    \hline
        Polynomial CBF \cite{peng2023safe} 
        & Explicit 
        & Smooth 
        & Distance
        & Convex %Conservative 
        & $1$
        & No
        & No\\
    \hline
        Duality-based CBF \cite{thirugnanam2022duality}
        & Implicit 
        & Nonsmooth
        & Distance
        & Convex%Conservative %Parameter-adjustable 
        & $1$
        & No%$\mathcal{C}$- Space 
        & No \\
    \hline
        Duality-based DCBF + MPC \cite{thirugnanam2022safety}
        & Implicit 
        & Nonsmooth
        & Distance
        & Nonconvex%Conservative %Parameter-adjustable 
        & $N$
        & No%$\mathcal{C}$- Space 
        & No \\
    \hline
        Minkowski-based CBF \cite{chen2025control}
        & Implicit 
        & Nonsmooth
        & Distance
        & Convex
        & $1$
        & No %$\mathcal{C}$- Space  
        & No \\  
    \hline
        \textbf{Our method}
        & Implicit  
        & Nonsmooth 
        & Supporting hyperplane
        & Convex
        & $N$
        & Yes %$\mathcal{C}$- Space  
        & Yes \\
    \hline
    \end{tabular}}
    % \vspace{-0.3cm}
    \label{table1}
\end{table*}

Works such as \cite{thirugnanam2022duality,thirugnanam2022safety,chen2025control} 
construct CBFs directly from the exact geometry of polytopes, avoiding smooth 
surrogate approximations. In particular, \cite{thirugnanam2022duality} employs the 
Minimum Distance Function (MDF) via duality-based optimization to build a 
nonsmooth CBF, while \cite{chen2025control} leverages Minkowski set operations 
to enable exact distance evaluation between polytopic sets. These approaches 
preserve geometric exactness and demonstrate safe traversal of narrow passages 
in 2-D simulations.
However, \cite{thirugnanam2022duality} and \cite{chen2025control} adopt 
single-step control updates without explicit look-ahead, which may lead to 
aggressive behavior in complex environments. In contrast, 
\cite{thirugnanam2022safety} combines DCBFs, duality-based distance computation, 
and Model Predictive Control (MPC) to incorporate future state information along a receding horizon, 
yielding safer control. Nevertheless, the resulting optimization remains 
nonconvex, and practical implementations typically rely on short horizons or 
nearby obstacle selection for real-time performance. Moreover, in 3-D, Minkowski operations 
substantially increase polytope complexity. As a result, 
despite their geometric exactness and strong 2-D performance, these methods 
have not demonstrated efficient obstacle avoidance for fully three-dimensional 
robotic systems. Table~I provides a conceptual comparison.

In this paper, we propose an iterative convex optimization framework for obstacle avoidance between polytopes using discrete-time CBF constraints. The proposed framework can serve as a local planner to generate dynamically feasible and collision-free trajectories, or be directly deployed as a safety-critical controller for general dynamical systems. In particular, the contributions are as follows:
\begin{itemize}
    \item We construct supporting hyperplanes from exact closest-point computations between polytopes, yielding linear DCBF constraints for collision avoidance. 
    We then develop a novel iterative MPC-DCBF framework that keeps the finite-horizon optimization convex at each iteration, enabling fast online computation for safety-critical control and planning of nonlinear systems.

    \item We extend the framework to multi-robot systems through a sequential scheme while preserving convex optimization structure. 

    \item We provide numerical validation in 2-D and 3-D maze environments with polytopic obstacles, where convex and nonconvex robots perform tight maneuvers through narrow passages and multi-robot interactions, while maintaining real-time computational performance for control and trajectory generation.
\end{itemize}

\section{Preliminaries}
\label{sec:Preliminaries}

In this section, we first review optimization-based formulations using Discrete-time High-order CBFs (DHOCBFs), and then introduce the notation for obstacle avoidance of polytopic sets by a polytopic robot.

\subsection{Optimization Formulation using DHOCBFs}

We consider a discrete-time control system in the form:
\begin{equation}
\label{eq:discrete-dynamics}
\mathbf{x}_{t+1} = f(\mathbf{x}_t, \mathbf{u}_t),
\end{equation}
where $\mathbf{x}_t \in \mathcal{X} \subset \mathbb{R}^n$ denotes the system state at time step $t \in \mathbb{N}$, $\mathbf{u}_t \in \mathcal{U} \subset \mathbb{R}^q$ is the control input, and the function $f: \mathbb{R}^n \times \mathbb{R}^q \to \mathbb{R}^n$ is assumed to be locally Lipschitz continuous.
Safety is defined in terms of forward invariance of a set $\mathcal{C}$. Specifically, system \eqref{eq:discrete-dynamics} is considered safe if, for any initial condition $\mathbf{x}_0 \in \mathcal{C}$, the state satisfies $\mathbf{x}_t \in \mathcal{C}$ for all $t \in \mathbb{N}$. We define $\mathcal{C}$ as the superlevel set of a continuous function $h: \mathbb{R}^n \to \mathbb{R}$:
\begin{equation}
\label{eq:safe-set}
\mathcal{C} \coloneqq \{ \mathbf{x} \in \mathbb{R}^n : h(\mathbf{x}) \geq 0 \}.
\end{equation}
\begin{definition}[Relative degree~\cite{liu2023iterative}]
\label{def:relative-degree}
The output $y_{t}=h(\mathbf{x}_{t})$ of system \eqref{eq:discrete-dynamics} is said to have relative degree $m$ if
\begin{equation}
\begin{split}
&y_{t+i}=h(\bar{f}_{i-1}(f(\mathbf{x}_{t},\mathbf{u}_{t}))), \ i \in \{1,2,\dots,m\},\\
 \text{s.t.} & \ \frac{\partial y_{t+m}}{\partial \mathbf{u}_{t}} \ne \textbf{0}_{q}, \frac{\partial y_{t+i}}{\partial \mathbf{u}_{t}}= \textbf{0}_{q},  \ i \in \{1,2,\dots,m-1\},
\end{split}
\end{equation}
i.e., $m$ is the number of steps (delay) in the output $y_{t}$ in order for any component of the control input $\mathbf{u}_{t}$ to explicitly appear ($\textbf{0}_{q}$ is the zero vector of dimension $q$). 
\end{definition}

In Def.~\ref{def:relative-degree}, $\bar{f}(\mathbf{x})\coloneqq f(\mathbf{x},0)$ denotes the uncontrolled state dynamics. 
The subscript of $\bar{f}$ indicates recursive composition:
$\bar{f}_{0}(\mathbf{x})=\mathbf{x}$ and 
$\bar{f}_{i}(\mathbf{x})=\bar{f}(\bar{f}_{i-1}(\mathbf{x}))$ for $i\ge1$.
We assume that $h(\mathbf{x})$ has
relative degree $m$ with respect to system (\ref{eq:discrete-dynamics}) based on Def. \ref{def:relative-degree}.
Starting with $\psi_{0}(\mathbf{x}_{t})\coloneqq h(\mathbf{x}_{t})$, we define a sequence of discrete-time functions $\psi_{i}:  \mathbb{R}^{n}\to\mathbb{R}$, $i=1,\dots,m$ as:
\begin{equation}
\label{eq:high-order-discrete-CBFs}
\psi_{i}(\mathbf{x}_{t})\coloneqq \bigtriangleup \psi_{i-1}(\mathbf{x}_{t})+\alpha_{i}(\psi_{i-1}(\mathbf{x}_{t})), 
\end{equation}
where $\bigtriangleup \psi_{i-1}(\mathbf{x}_{t})\coloneqq \psi_{i-1}(\mathbf{x}_{t+1})-\psi_{i-1}(\mathbf{x}_{t})$, and $\alpha_{i}(\cdot)$ denotes the $i^{th}$ class $\kappa$ function which satisfies $\alpha_{i}(\psi_{i-1}(\mathbf{x}_{t}))\le \psi_{i-1}(\mathbf{x}_{t})$ for $i=1,\ldots, m$.
A sequence of sets $\mathcal {C}_{i}$ is defined based on \eqref{eq:high-order-discrete-CBFs} as
\begin{equation}
\label{eq:high-order-safety-sets}
\mathcal {C}_{i}\coloneqq \{\mathbf{x}\in \mathbb{R}^{n}:\psi_{i}(\mathbf{x})\ge 0\}, \ i \in\{0,\ldots,m-1\}.
\end{equation}

\begin{definition}[DHOCBF~\cite{xiong2022discrete}]
\label{def:high-order-discrete-CBFs}
Let $\psi_{i}(\mathbf{x}), \ i\in \{1,\dots,m\}$ be defined by \eqref{eq:high-order-discrete-CBFs} and $\mathcal {C}_{i},\ i\in \{0,\dots,m-1\}$ be defined by \eqref{eq:high-order-safety-sets}. A function $h:\mathbb{R}^{n}\to\mathbb{R}$ is a Discrete-time High-Order Control Barrier Function (DHOCBF) with relative degree $m$ for system \eqref{eq:discrete-dynamics} if there exist $\psi_{m}(\mathbf{x})$ and $\mathcal {C}_{i}$ such that
\begin{equation}
\label{eq:highest-order-CBF}
\psi_{m}(\mathbf{x}_{t})\ge 0, \ \forall x_{t}\in \mathcal{C}_{0}\cap \dots \cap \mathcal {C}_{m-1}, t\in\mathbb{N}.
\end{equation}
\end{definition}

\begin{theorem}[Safety Guarantee \cite{xiong2022discrete}]
\label{thm:forward-invariance}
Given a DHOCBF $h(\mathbf{x})$ from Def. \ref{def:high-order-discrete-CBFs} with corresponding sets $\mathcal{C}_{0}, \dots,\mathcal {C}_{m-1}$ defined by \eqref{eq:high-order-safety-sets}, if $\mathbf{x}_{0} \in \mathcal {C}_{0}\cap \dots \cap \mathcal {C}_{m-1},$ then any Lipschitz controller $\mathbf{u}_{t}$ that satisfies the constraint in \eqref{eq:highest-order-CBF}, $\forall t\ge 0$ renders $\mathcal {C}_{0}\cap \dots \cap \mathcal {C}_{m-1}$ forward invariant for system \eqref{eq:discrete-dynamics}, $i.e., \mathbf{x}_{t} \in \mathcal {C}_{0}\cap \dots \cap \mathcal {C}_{m-1}, \forall t\ge 0.$
\end{theorem}
We can simply define an $i^{th}$ order DCBF $\psi_{i}(\mathbf{x})$ in \eqref{eq:high-order-discrete-CBFs} as
\begin{equation}
\label{eq:simple-high-order-discrete-CBFs}
\psi_{i}(\mathbf{x}_{t})\coloneqq \bigtriangleup \psi_{i-1}(\mathbf{x}_{t})+\gamma_{i}\psi_{i-1}(\mathbf{x}_{t}),
\end{equation}
where $\alpha(\cdot)$ is defined linear and $0<\gamma_{i}\le 1, i\in \{1,\dots,m\}$. Rewrite \eqref{eq:simple-high-order-discrete-CBFs}, we have $\psi_{i-1}(\mathbf{x}_{t+1})\ge (1-\gamma_{i})\psi_{i-1}(\mathbf{x}_{t}).$ This shows 
the lower bound of $\psi_{i-1}$ decreases exponentially with the rate $1-\gamma_{i}$.

\subsection{Obstacle Avoidance between Polytopic Sets}
Let the robot state (configuration) be $\mathbf{x} \in \mathbb{R}^n$ with discrete-time dynamics given in \eqref{eq:discrete-dynamics}. 
We consider the geometry of both the robot and the obstacles in an $\ell$-dimensional workspace. 
The $i_{o}$-th static obstacle and the robot at state $\mathbf{x} \in \mathcal{X}$ are modeled as convex polytopes:
\begin{align}
\label{eq:obs}
\mathcal{O}_{i_{o}} &:= \{ \mathbf{c} \in \mathbb{R}^\ell : A^{\mathcal{O}_{i_{o}}} \mathbf{c} \le b^{\mathcal{O}_{i_{o}}} \}, \\
\label{eq:rob}
\mathcal{R}(\mathbf{x}) &:= \{ \mathbf{c} \in \mathbb{R}^\ell : A^{\mathcal{R}}(\mathbf{x}) \mathbf{c} \le b^{\mathcal{R}}(\mathbf{x}) \},
\end{align}
where $b^{\mathcal{O}_{i_{o}}} \in \mathbb{R}^{s^{\mathcal{O}_{i_{o}}}}$ and ${i_{o}} \in \{1,\dots,N_{\mathcal{O}}\}$. 
The vector $\mathbf{c} \in \mathbb{R}^\ell$ denotes a point in the workspace. 
The matrices $A^{\mathcal{R}}(\mathbf{x})$ and vectors $b^{\mathcal{R}}(\mathbf{x}) \in \mathbb{R}^{s^{\mathcal{R}}}$ are assumed to be continuous in $\mathbf{x}$. 
Inequalities involving vectors are interpreted element-wise. 
We assume that $\mathcal{O}_{i_{o}}$, ${i_{o}} \in \{1,\dots,N_O\}$, and $\mathcal{R}(\mathbf{x})$ for all $\mathbf{x} \in \mathcal{X}$ are bounded and non-empty. 
The quantities $s^{\mathcal{O}_{i_{o}}}$ and $s^{\mathcal{R}}$ denote the number of facets of the polytopic sets corresponding to the ${i_{o}}$-th obstacle and the robot, respectively. For all ${i_{o}} \in \{1,\dots,N_O\}$ and all $x \in \mathcal{X}$, 
the sets $\mathcal{O}_{i_{o}}$ and $\mathcal{R}(\mathbf{x})$ are non-empty, convex, and compact. 
Consequently, the minimum distance between any pair $(\mathcal{O}_{i_{o}}, \mathcal{R}(\mathbf{x}))$ is well-defined. 
Moreover, this distance is zero if and only if $\mathcal{O}_{i_{o}}$ and $\mathcal{R}(\mathbf{x})$ intersect.

The square of the minimum distance between $\mathcal{O}_{i_{o}}$ and $\mathcal{R}(\mathbf{x})$, 
denoted by $h_{i_{o}}(\mathbf{x})$, can be computed via the following:
\begin{align}
&h_{i_{o}}(\mathbf{x}) = \min_{\mathbf{c}^{\mathcal{O}_{i_{o}}}\in \mathbb{R}^{\ell},\, \mathbf{c}^{\mathcal{R}} \in \mathbb{R}^{\ell}}
\| \mathbf{c}^{\mathcal{O}_{i_{o}}} - \mathbf{c}^{\mathcal{R}} \|_2^2 \label{eq:dist_qp_obj} \\
\text{s.t.} \quad 
& A^{\mathcal{O}_{i_{o}}} \mathbf{c}^{\mathcal{O}_{i_{o}}} \le b^{\mathcal{O}_{i_{o}}}, ~
 A^{\mathcal{R}}(\mathbf{x}) \mathbf{c}^{\mathcal{R}} \le b^{\mathcal{R}}(\mathbf{x}). \label{eq:dist_qp_con}
\end{align}
To ensure safe motion of the robot, a natural approach is to enforce the DHOCBF constraints~\eqref{eq:highest-order-CBF} pairwise for each robot–obstacle pair so as to guarantee forward invariance of the safe set $\mathcal{C}_{i_o}:=\{\mathbf{x}\mid h_{i_o}(\mathbf{x})\ge0\}$.
However, since $h_{i_o}(\mathbf{x})$ is generally nonlinear in $\mathbf{x}$, the resulting constraints~\eqref{eq:highest-order-CBF} become nonlinear, which may render the corresponding optimization problem nonconvex and increase computational complexity. 
We address this issue in Sec.~\ref{sec:Iterative MPC}.

\section{Problem Formulation and Approach}
\label{sec:Problem Formulation and Approach}

The technical discussion in this paper is on a single controlled robot operating in an environment with static obstacles. An extension to multi-robot teams is presented at the end. The robot and all obstacles (or their over-approximations) are bounded polyhedra represented as unions of convex polytopes. 
The robot must reach a specified target while ensuring safety and respecting its dynamics and input constraints. 
The environment contains $N_o$ static obstacles 
$\mathcal{O}_1, \dots, \mathcal{O}_{N_o} \subset \mathbb{R}^\ell$, $\ell \in \{2,3\}$, each with known and fixed geometric boundaries. Based on Eqs. \eqref{eq:obs}--\eqref{eq:rob}, let 
$\mathcal{O} \coloneqq \bigcup_{i_o=1}^{N_o} \mathcal{O}_{i_o}$
denote the union of all static obstacles. Let 
$\mathcal{R}(\mathbf{x}_t) \subset \mathbb{R}^\ell$
denote the occupied region of the robot at time $t$, 
which depends on its configuration $\mathbf{x}_t$. 
The safety requirement for the robot is then written as
\begin{equation}
\label{eq: safety}
\mathcal{R}(\mathbf{x}_t) \cap \mathcal{O} = \emptyset,
\quad \forall t \ge 0.
\end{equation}

We assume that the robot dynamics are described by \eqref{eq:discrete-dynamics} and that the obstacles are represented in the configuration space (C-space). In this paper, we consider the following problem:
\begin{figure}
\vspace{3mm}
    \centering
    \includegraphics[scale=0.22]{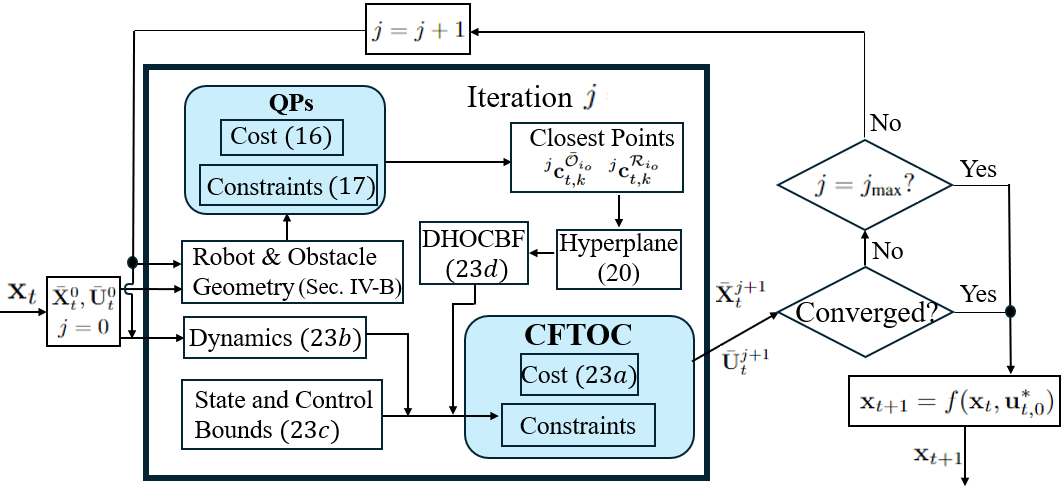}
    \caption{Schematic of the iterative process of solving the convex MPC at time $t$.}
    \label{fig:iteration-module}
\end{figure}
\begin{problem}
\label{prob:Path-prob}
For a robot with geometry $\mathcal{R}$ and dynamics~\eqref{eq:discrete-dynamics}, 
moving in an environment with obstacles $\mathcal{O}_{i_o}$, $i_o=1,\ldots,N_o$, 
design control inputs that steer the robot from its initial configuration 
$\mathbf{x}_0$ to a circular target region centered at $\mathbf{x}_T$ in finite time, 
while ensuring the safety condition~\eqref{eq: safety}, 
satisfying the state and input constraints 
$\mathbf{x}_t\in\mathcal{X}\subset\mathbb{R}^n$, 
$\mathbf{u}_t\in\mathcal{U}\subset\mathbb{R}^q$, 
and minimizing the control effort 
$\sum_{t=0}^{T-1}\|\mathbf{u}_t\|^2$.
\end{problem}

In the case of a single robot operating in a 
two-dimensional environment ($\ell=2$), 
\cite{thirugnanam2022safety} combines DCBFs, duality-based distance computation, 
and MPC to achieve obstacle avoidance. 
However, the resulting optimization problem is nonconvex, 
leading to high computational complexity. To mitigate this issue, several practical simplifications are introduced, 
such as using dual variables, 
shortening the prediction horizon, and restricting the scenario to a 
single robot in a 2-D setting. 
While these strategies improve computational tractability, 
they limit the scalability and broader applicability of the method. 

\textbf{Approach:} We address Problem~\ref{prob:Path-prob} within an MPC framework 
while ensuring that the resulting optimization problem remains convex. 
The MPC formulation enforces linear constraints on the decision variables, 
including the system dynamics, state and input constraints, and DHOCBF 
safety constraints, while minimizing the quadratic control cost defined 
in Problem~\ref{prob:Path-prob}. 
At each time step, the resulting convex MPC problem is solved iteratively 
until the predicted trajectories converge.

To obtain linear dynamics constraints, we linearize the nonlinear system 
dynamics in \eqref{eq:discrete-dynamics} around the nominal states. To render the DHOCBF constraints linear, supporting hyperplanes are constructed using \eqref{eq:dist_qp_obj} and \eqref{eq:dist_qp_con}. 
The relative geometry $\mathbf{c}^{\mathcal{O}_{i_o}} - \mathbf{c}^{\mathcal{R}}$ from the closest-point solution defines the hyperplane normal, yielding linear DHOCBF constraints in the decision variables. This construction leads to a convex MPC problem that can be solved 
efficiently, avoiding the nonconvex coupling present in previous 
approaches such as~\cite{thirugnanam2022safety}.
Although the formulation focuses on a single robot navigating among static obstacles, 
the same construction naturally extends to multi-robot systems by treating 
other robots as dynamic polytopic obstacles. 
For this case, we present a sequential scheme can be employed while preserving convexity of optimization.

\section{Iterative MPC with DHOCBF constraints}
\label{sec:Iterative MPC}
\label{sec:Methodology}

Our method is implemented within a receding-horizon framework. 
At each time step $t$, the prediction horizon involves the state sequence 
$\mathbf{x}_{t,k},~ k \in \{0,\dots,N\}$ 
and the input sequence 
$\mathbf{u}_{t,k},~ k \in \{0,\dots,N-1\}$. 
Fig.~\ref{fig:iteration-module} illustrates the iterative solution procedure at time $t$.

The optimal solution obtained at the previous time step, 
$\mathbf{X}_{t-1}^{*}$ and $\mathbf{U}_{t-1}^{*}$, 
is used to initialize the nominal trajectory for iteration $j=0$ at time $t$, i.e., 
$\bar{\mathbf{X}}_{t}^{0} = \mathbf{X}_{t-1}^{*}, 
~
\bar{\mathbf{U}}_{t}^{0} = \mathbf{U}_{t-1}^{*}.$
For $t=0$, the initial nominal input $\bar{\mathbf{U}}_{0}^{0}$ is designed such that 
the corresponding nominal state $\bar{\mathbf{X}}_{0}^{0}$ satisfies the safety requirements. At each iteration $j$, auxiliary QPs are first solved to compute the closest points and the associated supporting hyperplanes. 
Subsequently, a convex finite-time optimal control (CFTOC) problem is solved with linearized dynamics and linear DHOCBF constraints, yielding
$\mathbf{X}_{t}^{*,j} = [\mathbf{x}_{t,0}^{*,j},\dots,\mathbf{x}_{t,N}^{*,j}], 
~
\mathbf{U}_{t}^{*,j} = [\mathbf{u}_{t,0}^{*,j},\dots,\mathbf{u}_{t,N-1}^{*,j}].$
The nominal trajectories are then updated as
$\bar{\mathbf{X}}_{t}^{j+1} = \mathbf{X}_{t}^{*,j}, 
~ \bar{\mathbf{U}}_{t}^{j+1} = \mathbf{U}_{t}^{*,j}.$
The iterative procedure terminates when a prescribed normalized convergence tolerance is satisfied or when the maximum number of iterations $j_{\max}$ is reached. 
The resulting optimal trajectories $\mathbf{X}_{t}^{*}$ and $\mathbf{U}_{t}^{*}$ are used at the next time step.
Note that the open-loop nominal trajectories 
$\bar{\mathbf{X}}_{t}^{j}$ and $\bar{\mathbf{U}}_{t}^{j}$ 
are propagated between iterations, enabling successive local linearization of the system dynamics. 
Finally, the closed-loop system evolves according to
$\mathbf{x}_{t+1} = f(\mathbf{x}_{t}, \mathbf{u}_{t,0}^{*}).$

\subsection{Linearization of Dynamics}
\label{subsec:L of dynamics}
At iteration $j$, a control vector $\mathbf{u}_{t,k}^{j}$ is obtained by linearizing the system around the nominal pair $(\bar{\mathbf{x}}_{t,k}^{j}, \bar{\mathbf{u}}_{t,k}^{j})$:
\begin{equation}
\begin{split}
\label{eq:linearized-dynamics}
\mathbf{x}_{t,k+1}^{j}-\bar{\mathbf{x}}_{t,k+1}^{j}
=
A_{t,k}^{j}\big(\mathbf{x}_{t,k}^{j}-\bar{\mathbf{x}}_{t,k}^{j}\big)
+\\
B_{t,k}^{j}\big(\mathbf{u}_{t,k}^{j}-\bar{\mathbf{u}}_{t,k}^{j}\big)
+
d_{t,k}^{j},
\end{split}
\end{equation}
where $d_{t,k}^{j}=f(\bar{\mathbf{x}}_{t,k}^{j}, \bar{\mathbf{u}}_{t,k}^{j})-\bar{\mathbf{x}}_{t,k+1}^{j}$, $0 \leq j < j_{\max}$, and $k$ and $j$ denote the open-loop time-step index and the iteration index, respectively. Moreover,
\begin{equation}
A_{t,k}^{j}=D_{\mathbf{x}}f(\bar{\mathbf{x}}_{t,k}^{j}, \bar{\mathbf{u}}_{t,k}^{j}),
~
B_{t,k}^{j}=D_{\mathbf{u}}f(\bar{\mathbf{x}}_{t,k}^{j}, \bar{\mathbf{u}}_{t,k}^{j}),
\end{equation}
where $D_{\mathbf{x}}$ and $D_{\mathbf{u}}$ denote the Jacobians of the dynamics \eqref{eq:discrete-dynamics} with respect to the state $\mathbf{x}$ and the input $\mathbf{u}$, respectively.
This linearization is performed locally at $(\bar{\mathbf{x}}_{t,k}^{j}, \bar{\mathbf{u}}_{t,k}^{j})$ and updated between iterations. In iteration $j$, the nominal vectors $(\bar{\mathbf{x}}_{t,k}^{j}, \bar{\mathbf{u}}_{t,k}^{j})$ are treated as constants (constructed from iteration $j{-}1$), and thus the dynamics constraint in \eqref{eq:linearized-dynamics} is linear.

\subsection{Robot and Obstacle Geometry}
\label{subsec: L of geometry}
The nonconvexity in \eqref{eq:dist_qp_con} arises from the state-dependent
robot geometry described by 
$A^{\mathcal R}(\mathbf{x})$ and $b^{\mathcal R}(\mathbf{x})$.
To preserve convexity within each iteration, we evaluate 
$A^{\mathcal R}(\mathbf{x})$ and $b^{\mathcal R}(\mathbf{x})$ 
at the nominal state and treat them as constants, 
which renders the constraints \eqref{eq:dist_qp_con} linear in the decision variables.

Specifically, for each nominal state $\bar{\mathbf{x}}_{t,k}^{j}$, we consider only polygonal obstacles within an axis-aligned box.
The number of such obstacles is defined as
\begin{equation}
\begin{aligned}
\bar{N}_{t,k}^{j}
=
\Big|
\{\, 
i_o \in \{1,\dots,N_{\mathcal O}\} :
\exists\,\mathbf c^{\mathcal{O}_{i_{o}}}  \in \mathcal O_{i_o} \\
\text{ s.t. }
\|\mathbf c^{\mathcal{O}_{i_{o}}}  - \bar{\mathbf x}_{t,k}^j\|_\infty\le r_o
\}
\,\Big|.
\end{aligned}
\end{equation}
Here $|\cdot|$ denotes the cardinality of a set, 
$\mathbf c^{\mathcal O_{i_o}}$ represents a point belonging to the obstacle 
$\mathcal O_{i_o}$, and $\|\cdot\|_\infty$ denotes the infinity norm, 
which defines an axis-aligned square (in 2D) or cube (in 3D). The collection of polygonal obstacles within this axis-aligned box is denoted by $\bar{\mathcal O}$. For each $i_o \in \{0,\dots,\bar{N}_{t,k}^j\}$, we solve the following QP to compute the corresponding pair of closest points between the robot set and the $i_o$-th active obstacle:
\begin{align}
 &\min_{{}^{j}\mathbf{c}_{t,k}^{\bar{\mathcal 
 O}_{i_o}}\in \mathbb{R}^{\ell},\,
      {}^{j}\mathbf{c}_{t,k}^{\mathcal R_{i_o}} \in \mathbb{R}^{\ell}}
  \left\|
 {}^{j}\mathbf{c}_{t,k}^{\bar{\mathcal O}_{i_o}}
 -
 {}^{j}\mathbf{c}_{t,k}^{\mathcal R_{i_o}}
 \right\|_2^2
 \label{eq:dist_qp_obj2} \\
\text{s.t.} \quad 
& A^{\bar{\mathcal O}_{i_o}}
{}^{j}\mathbf{c}_{t,k}^{\bar{\mathcal O}_{i_o}}
\le b^{\bar{\mathcal O}_{i_o}}, ~
 A^{\mathcal R}(\bar{\mathbf {x}}_{t,k}^{j})
{}^{j}\mathbf{c}_{t,k}^{\mathcal R_{i_o}}
\le b^{\mathcal R}(\bar{\mathbf {x}}_{t,k}^{j}).
\label{eq:dist_qp_con2}
\end{align}
The optimal solutions 
${}^{j}\mathbf{c}_{t,k}^{\bar{\mathcal O}_{i_o}}$ 
and 
${}^{j}\mathbf{c}_{t,k}^{\mathcal R_{i_o}}$ 
denote the closest points on the $i_o$-th obstacle and the robot set, respectively. 
Since each robot-side closest point corresponds uniquely to a specific obstacle, 
the robot variable is also indexed by $i_o$. If the robot and obstacle share parallel supporting faces, the closest-point pair may not be unique. 
In this case, any optimal solution of the QP is sufficient, as all such solutions yield the same minimum distance.

\subsection{Supporting Hyperplanes}
\label{subsec: SH}
Having obtained the closest-point pair 
${}^{j}\mathbf c_{t,k}^{\bar{\mathcal O}_{i_o}}$ 
and 
${}^{j}\mathbf c_{t,k}^{\mathcal R_{i_o}}$ 
from the QP, 
we next exploit their geometric structure. 
In particular, the vector connecting the two closest points 
naturally induces a separating direction between the robot set 
and the obstacle. 
The following proposition formalizes this property.
\begin{proposition}
\label{lem:hyperplane}
Let $\mathcal R$ and $\bar{\mathcal O}_{i_o}$ be two disjoint convex polytopes,
and let 
${}^{j}\mathbf c_{t,k}^{\mathcal R_{i_o}}$ 
and 
${}^{j}\mathbf c_{t,k}^{\bar{\mathcal O}_{i_o}}$
be a closest-point pair. Assume $\mathcal R \cap \bar{\mathcal O}_{i_o} = \emptyset$,
then the hyperplanes orthogonal to $\mathbf{n}_{t,k}^{j,i_o}
=
{}^{j}\mathbf c_{t,k}^{\mathcal R_{i_o}}
-
{}^{j}\mathbf c_{t,k}^{\bar{\mathcal O}_{i_o}}$
and passing through the respective closest points
are supporting hyperplanes of the two polytopes
and separate them.
\end{proposition}
The proof follows from standard results in convex analysis and is omitted.

\subsection{Iterative MPC with DHOCBF Constraints}
\label{subsec:iMPC DHOCBFs}
Let the robot state be $\mathbf x = [\mathbf{p}^\top, \boldsymbol\theta^\top]^\top$, 
where $\mathbf{p}\in\mathbb R^\ell$ ($\ell=2$ or $3$) denotes the robot position and $\boldsymbol\theta\in\mathbb R^{\ell_\theta}$ 
parameterizes the robot orientation (e.g., $\ell_\theta=1$ in 2D and $\ell_\theta\in\{2,3\}$ in 3D).
Any point on the robot polytope can be written as
\begin{equation}
\mathbf{c}_{\mathrm{robot}}(\mathbf x)
= \mathbf{p} + R(\boldsymbol\theta)\, \mathbf{c}_{\mathrm{local}},
\label{eq:y_robot}
\end{equation}
where $R(\boldsymbol\theta)$ denotes the rotation matrix associated with the orientation $\boldsymbol\theta$, and $
\mathbf c_{\mathrm{local}}
=
R(\bar{\boldsymbol\theta})^\top
\big(\bar{\mathbf c}^{\mathcal R_{i_o}}
-
\bar{\mathbf{p}}
\big)$
is the corresponding closest point expressed in the robot body frame at the nominal state $\bar{\mathbf{x}}=(\bar{\mathbf{p}},\bar{\boldsymbol\theta})$. $\bar{\mathbf c}^{\mathcal R_{i_o}}$ denotes the robot-side closest point obtained from the distance QP formulated in Sec. \ref{subsec: L of geometry}. Since $R(\boldsymbol\theta)$ depends nonlinearly on the orientation, 
we linearize \eqref{eq:y_robot} about $\bar{\boldsymbol\theta}$ to obtain

\begin{equation}
\label{eq: L of crobot}
\mathbf{c}_{\mathrm{robot}}(\mathbf x)
\approx
\mathbf{p} + R(\bar{\boldsymbol\theta})\, \mathbf{c}_{\mathrm{local}}
+\sum_{r=1}^{\ell_\theta}
(\theta_{r}-\bar{\theta}_r)\,
\left.\frac{\partial R(\boldsymbol\theta)}{\partial \theta_r}\right|_{\boldsymbol\theta=\bar{\boldsymbol\theta}}
\mathbf{c}_{\mathrm{local}}.
\end{equation}

$\mathbf c_{\mathrm{robot}}(\mathbf x)$ represents the robot-side closest point 
corresponding to state $\mathbf x$. 
From \eqref{eq: L of crobot}, it can be seen that computing 
$\mathbf c_{\mathrm{robot}}(\mathbf x)$ requires the nominal state 
$\bar{\mathbf x}$. 
Therefore, at time $t+k$ and iteration $j$, 
$\mathbf c_{\mathrm{robot}}(\mathbf x)$ should be written as $\mathbf c_{\mathrm{robot}}(
\mathbf x_{t,k}^{j}
\mid
\bar{\mathbf x}_{t,k}^{j}
).$
From Sec. \ref{subsec: SH}, we obtain the outward normal direction 
$\mathbf{n}_{t,k}^{j,i_o}$ pointing away from the obstacle. 
The supporting hyperplane passing through the obstacle-side closest point 
${}^{j}\mathbf c_{t,k}^{\bar{\mathcal O}_{i_o}}$ 
and orthogonal to $\mathbf{n}_{t,k}^{j,i_o}$ is given by
\begin{equation}
\label{eq: hyperplane}
(\mathbf{n}_{t,k}^{j,i_o})^\top 
\big(\mathbf{x} - {}^{j}\mathbf c_{t,k}^{\bar{\mathcal O}_{i_o}}\big) = 0.
\end{equation}
According to Proposition~\ref{lem:hyperplane}, this hyperplane separates the robot and the obstacle, 
thereby preventing intersection of the two polytopes. In order to guarantee safety with forward invariance based on Thm. \ref{thm:forward-invariance} and Eq. \eqref{eq: hyperplane}, we derive a sequence of DHOCBFs up to the order $m$:
\begin{equation}
\label{eq:DCBFs}
\begin{split}
& \tilde{\psi}_{0}^{i_{o}}(\mathbf{x}_{t,k}^{j}) \coloneqq  (\mathbf{n}_{t,k}^{j,i_o})^\top 
\big(\mathbf c_{\mathrm{robot}}(
\mathbf x_{t,k}^{j}
\mid
\bar{\mathbf x}_{t,k}^{j}
) - {}^{j}\mathbf c_{t,k}^{\bar{\mathcal O}_{i_o}}\big),\\&
 \tilde{\psi}_{i}^{i_{o}}(\mathbf{x}_{t,k}^{j}) \coloneqq \tilde{\psi}_{i-1}^{i_{o}}(\mathbf{x}_{t,k+1}^{j})-\tilde{\psi}_{i-1}^{i_{o}}(\mathbf{x}_{t,k}^{j}){+}\gamma_{i}^{i_{o}}\tilde{\psi}_{i-1}^{i_{o}}(\mathbf{x}_{t,k}^{j}),
 \end{split}
\end{equation}
where $0 <\gamma_{i}^{i_{o}} \le 1$, $i\in\{1,\dots,m\}, i_{o}\in \{0,\dots,\bar{N}_{t,k}^j\}.$ Note that as the prediction horizon $N$, the relative degree $m$, and the number of detected obstacles $\bar{N}_{t,k}^j$ increase, the number of constraints at each iteration may grow significantly, potentially rendering the optimization problem infeasible. 
To address this issue, we introduce a slack variable $\omega_{t,k,i}^{j,i_{o}}$ with an associated decay rate $(1-\gamma_{i}^{i_o})$. Similar to \cite{liu2025learning}, we replace $\tilde{\psi}_{i}^{i_{o}}(\mathbf{x}_{t,k}^{j})\ge 0$ in \eqref{eq:DCBFs} with
\begin{equation}
%{\small
\label{eq:convex-hocbf-constraint}
\begin{split}
 \tilde{\psi}_{i-1}^{i_{o}}(\mathbf{x}_{t,k}^{j})& + \sum_{\nu =1}^{i}(\gamma_{i}^{i_{o}}-1)Z_{\nu,i}^{i_{o}}(1-\gamma_{1}^{i_{o}})^{k-1}\tilde{\psi}_{0}^{i_{o}}(\mathbf{x}_{t,\nu}^{j})\ge \\
\omega_{t,k,i}^{j,i_{o}}&(1-\gamma_{i}^{i_{o}}) Z_{0,i}^{i_{o}}(1-\gamma_{1}^{i_{o}})^{k-1}\tilde{\psi}_{0}^{i_{o}}(\mathbf{x}_{t,0}^{j}), 
  % i_{o}\in& \{0,\dots,\bar{N}_{t,k}^j\}, \ i\in \{1,\dots,m\},
\end{split}
\end{equation}
where $Z_{\nu,i}^{i_{o}}$ is a constant that aims to make constraint \eqref{eq:convex-hocbf-constraint} linear in terms of decision variables $\mathbf{x}_{t,k}^{j}, \omega_{t,k,i}^{j,i_{o}},$ and can be obtained in [Eq. (21), \cite{liu2025learning}].
The admissible range of $\omega_{t,k,i}^{j,i_o}$ can be determined from [Eq. (20), \cite{liu2025learning}]
to enlarge feasibility without compromising safety.

The linearization of the system dynamics and the construction 
of supporting hyperplanes as linear DHOCBF constraints enable the incorporation of safety constraints into a convex MPC formulation 
at each iteration, which we refer to as convex finite-time constrained optimal control (CFTOC). At iteration $j$, the CFTOC problem is solved with optimization variables $\mathbf U_t^j = [\mathbf u_{t,0}^j, \dots, \mathbf u_{t,N-1}^j]$
and $
\Omega_{t,i}^{j,i_{o}} = [\omega_{t,1,i}^{j,i_{o}}, \dots, \omega_{t,N,i}^{j,i_{o}}],$
where $i \in \{1,\dots,m\}$ and 
$i_{o} \in \{0,\dots,\bar{N}_{t,k}^j\}$.

\noindent\rule{\columnwidth}{0.4pt}
  \textbf{CFTOC of iMPC-DHOCBF at iteration $j$:}  
\begin{subequations}
{\small
\label{eq:impc-dcbf}
\begin{align}
\label{eq:impc-dcbf-cost}
  & \min_{\mathbf{U}_{t}^{j},\mathbf \Omega_{t,1}^{j},\dots, \mathbf \Omega_{t,m}^{j}} p(\mathbf{x}_{t,N}^{j})+\sum_{k=0}^{N-1} q(\mathbf{x}_{t,k}^{j},\mathbf{u}_{t,k}^{j},\omega_{t,k,i}^{j,l}) \\
   \text{s.t.} \ & \mathbf{x}_{t,k+1}^{j}{-} \bar{\mathbf{x}}_{t,k+1}^{j}{=}A^{j}(\mathbf{x}_{t,k}^{j}-\bar{\mathbf{x}}_{t,k}^{j}){+}B^{j}(\mathbf{u}_{t,k}^{j}-\bar{\mathbf{u}}_{t,k}^{j})+d_{t,k}^{j}, \label{subeq:impc-dcbf-linearized-dynamics}\\
    & \mathbf{u}_{t,k}^{j} \in \mathcal U, \  \mathbf{x}_{t,k}^{j} \in \mathcal X, \ \omega_{t,k,i}^{j,l}~ \text{s.t.} ~[\text{Eq}. (20), [12]],\label{subeq:impc-dcbf-variables-bounds}\\
    &  \tilde{\psi}_{i-1}^{i_{o}}(\mathbf{x}_{t,k}^{j}) + \sum_{\nu =1}^{i}(\gamma_{i}^{i_{o}}-1)Z_{\nu,i}^{i_{o}}(1-\gamma_{1}^{i_{o}})^{k-1}\tilde{\psi}_{0}^{i_{o}}(\mathbf{x}_{t,\nu}^{j})\ge \nonumber\\&
\omega_{t,k,i}^{j,i_{o}}(1-\gamma_{i}^{i_{o}}) Z_{0,i}^{i_{o}}(1-\gamma_{1}^{i_{o}})^{k-1}\tilde{\psi}_{0}^{i_{o}}(\mathbf{x}_{t,0}^{j}), \label{subeq:impc-dcbf-linearized-hocbf}
\end{align}
}
\end{subequations}
\noindent\rule{\columnwidth}{0.4pt}
In the CFTOC formulation, the linearized dynamics~\eqref{eq:linearized-dynamics} 
and the DHOCBF constraints~\eqref{eq:convex-hocbf-constraint} 
are enforced through~\eqref{subeq:impc-dcbf-linearized-dynamics} 
and~\eqref{subeq:impc-dcbf-linearized-hocbf}, respectively, 
for each open-loop step $k \in \{0,\dots,N-1\}$. 
State and input bounds are incorporated via~\eqref{subeq:impc-dcbf-variables-bounds}. 
The slack variables are assigned a reference value of $1$ in the cost function, 
encouraging minimal deviation from the original (unrelaxed) DHOCBF constraints and 
allowing relaxation only when necessary to maintain feasibility. 
To preserve the safety guarantees of the DHOCBF framework, 
constraints~\eqref{subeq:impc-dcbf-linearized-hocbf} 
are enforced for all $i \in \{1,\dots,m\}$ and 
$i_{o} \in \{0,\dots,\bar{N}_{t,k}^j\}$.
At iteration $j$, the optimal decision variables consist of the control input sequence
$\mathbf U_t^{*,j}=[\mathbf u_{t,0}^{*,j},\dots,\mathbf u_{t,N-1}^{*,j}]$
and the slack sequences
$\mathbf \Omega_{t,i}^{*,j}
=
[\Omega_{t,i}^{*,j,1},\dots,\Omega_{t,i}^{*,j,\bar N_{t,k}^j}],$
where
$\Omega_{t,i}^{*,j,i_o}
=
[\omega_{t,1,i}^{*,j,i_o},\dots,\omega_{t,N,i}^{*,j,i_o}]$.
The CFTOC problem is solved iteratively within the proposed iMPC-DHOCBF framework, 
where the dynamics and constraints are re-linearized at each iteration to reduce 
linearization errors; see~\cite{liu2025learning} for additional details.
\begin{remark}[Safety Margin under Geometry Linearization]
\label{rem:safety margin} 
At time $t$, we first solve the QPs in~\eqref{eq:dist_qp_obj2}--\eqref{eq:dist_qp_con2} 
to compute the closest-point pairs between the robot and obstacles, from which the supporting hyperplanes are constructed. 
These hyperplanes are then incorporated into the CFTOC problem~\eqref{eq:impc-dcbf} 
to obtain the state at time $t+1$. 
This sequential scheme avoids the nonlinear coupling between the state and closest-point variables and can be interpreted as a linearization strategy.
Approximation errors may arise from the time interval between closest-point evaluation and state update, as well as the geometric linearization in~\eqref{eq: L of crobot}. 
To mitigate these effects, a safety margin can be added to the hyperplane constraint~\eqref{eq: hyperplane} as 
$(\mathbf{n}_{t,k}^{j,i_o})^\top (\mathbf{x} - {}^{j}\mathbf c_{t,k}^{\bar{\mathcal O}_{i_o}}) \ge \epsilon$, 
where $\epsilon>0$ is a small constant. 
In the case studies, we set $\epsilon=0$ to evaluate the safety performance without additional conservatism.
\end{remark}
\subsection{Sequential Multi-Robot Implementation}
\label{subsec:decentralized control} 
Consider a team of $N_{r}$ robots indexed by $i_{r}=\{1,\ldots,N_{r}\}$. 
To avoid solving a large centralized optimization problem over all robots, we adopt a sequential scheme. At each time step $t$ and iteration $j$, robots are assigned a fixed priority ordering according to their indices. 
Robot $1$ first solves its MPC problem \eqref{eq:impc-dcbf} considering only static obstacles. 
After solving the optimization, it broadcasts its predicted trajectory over the MPC horizon.
For each subsequent robot $i_{r} \in \{2,\ldots,N_{r}\}$, the predicted trajectories of robots $1,\ldots,i_{r}-1$ are treated as known time-varying obstacles. 
Consequently, the corresponding robot--robot collision avoidance constraints are incorporated in the MPC problem of robot $i_{r}$ while remaining affine in its decision variables.
This procedure is repeated sequentially for all robots at each time step $t$ and iteration $j$. 
The resulting scheme requires only the exchange of predicted trajectories and avoids the need for a large optimization over all robots.

\subsection{Complexity Analysis}
\label{subsec: Complexity Analysis}
At each time step $t$ and iteration $j$, the proposed framework involves 
(i) solving a set of closest-point QPs for the detected obstacles and 
(ii) solving one CFTOC problem.

Let $\bar N_{t,k}^{j}$ denote the number of detected obstacles within the sensing range. 
Each closest-point computation is a QP with decision dimension $\mathcal O(\ell)$, where $\ell\in\{2,3\}$ is the workspace dimension. 
Thus, the geometric stage scales linearly with $\bar N_{t,k}^{j}$. 
The CFTOC is a convex QP with decision dimension $\mathcal O(N q + N m \bar N_{t,k}^{j})$, where $N$ is the prediction horizon, $q$ is the input dimension, and $m$ is the relative degree of the DHOCBF constraints. 
Using an interior-point method, the worst-case per-iteration complexity grows polynomially with the problem size. 
According to Remark~1 in~\cite{liu2025learning}, the highest DHOCBF order in~\eqref{eq:impc-dcbf} can be selected as $m_{\text{cbf}} \le m$. 
When $N$ is sufficiently large, the input $u$ appears in $\tilde{\psi}_{m_{\text{cbf}}}^{i_o}$ even for reduced order $m_{\text{cbf}}$, so choosing a smaller $m_{\text{cbf}}$ effectively lowers the computational burden.
In the multi-robot case with $N_r$ robots, the decentralized sequential scheme described earlier solves one CFTOC problem per robot while treating previously solved robots as dynamic obstacles. 
As a result, the total computational complexity scales approximately linearly with the number of robots.

\section{NUMERICAL RESULTS}
\label{sec: Numerical results}
We first focus on autonomous navigation for one robot of various shapes (rectangle, triangle, and L-shape) in 2-D environments. We then show extensions to multi-robot teams and 3-D environments. The proposed optimization-based control algorithm generates dynamically feasible, collision-free trajectories in tight maze scenarios. The animation video can be found at \url{https://youtu.be/6XmuV3Gxvm0}

All simulations were conducted using the same computational setup. 
The proposed iMPC-DHOCBF framework was implemented in Python, and all convex optimization problems were solved using OSQP~\cite{stellato2020osqp}. 
Experiments were performed on a Linux desktop with an Intel(R) Core(TM) i9-12900K CPU running at 3.20 GHz.

\subsection{One robot in a 2-D environment}
\label{subsec: 2-D}
\subsubsection{System Dynamics}
Consider a discrete-time unicycle model in the form
\begin{small}
\begin{equation}
\label{eq:unicycle-model}
\begin{bmatrix} x_{t+1}{-}x_t \\ y_{t+1}{-}y_t \\ \theta_{t+1}{-}\theta_t \\ v_{t+1}{-}v_t \end{bmatrix}{=}\begin{bmatrix} v_{t} \cos(\theta_{t}) \Delta t \\ v_{t}\sin(\theta_{t}) \Delta t \\ 0 \\ 0 \end{bmatrix}{+}\begin{bmatrix} 0 & 0 \\ 0 & 0 \\ \Delta t & 0 \\ 0 & \Delta t \end{bmatrix}
\begin{bmatrix} u_{1,t} \\ u_{2,t} \end{bmatrix},
\end{equation}
\end{small}
where $\mathbf{x}_{t}=[x_{t},y_{t},\theta_{t},v_{t}]^{\top}$ captures the 2-D location, heading angle, and linear speed; $\mathbf{u}_{t}=[u_{1,t},u_{2,t}]^{\top}$ represents angular velocity ($u_{1}$) and linear acceleration ($u_{2}$), respectively.
The system is discretized with $\Delta t = 0.1$.
System~\eqref{eq:unicycle-model} is subject to the following state and input constraints:
\begin{small}
\begin{equation}
\begin{split}
\label{eq:state-input-constraint}
\mathcal{X}&=\{\mathbf{x}_{t}\in \mathbb{R}^{4}: -2\cdot \mathcal{I}_{4\times1} \le \mathbf{x}_{t}\le 2\cdot \mathcal{I}_{4\times1}\},\\
\mathcal{U}&=\{\mathbf{u}_{t}\in \mathbb{R}^{2}: -0.5\cdot \mathcal{I}_{2\times1} \le \mathbf{u}_{t}\le 0.5\cdot \mathcal{I}_{2\times1}\}.
\end{split}
\end{equation}
\end{small}
\subsubsection{System Configuration}
The rectangle robot has a length of $0.15\,\mathrm{m}$ and a width of $0.06\,\mathrm{m}$, 
with its reference point located on the longitudinal centerline and offset $0.025\,\mathrm{m}$ from the rear edge. 
The triangle robot has an overall forward length of approximately $0.13\,\mathrm{m}$ and a rear width of $0.075\,\mathrm{m}$, 
with the reference point approximately at its centroid. 
The L-shape robot consists of two perpendicular slender legs forming a right angle, 
each with an overall extent of approximately $0.17\,\mathrm{m}$ and a thickness of $0.08\,\mathrm{m}$, 
and its reference point located at the inner corner where the two legs meet.
The initial states are $\mathbf{x}_0 = [0.15,\;0.225,\;0.0,\;0.0]^\top$, and aim to reach the target position $\mathbf{x}_T = [1.275,\;0.975]^\top$. The other reference vectors are $\mathbf{u}_{r} = [0,0]^{\top }$ and $\Omega_{r} = [1,1]^{\top }$.
\subsubsection{Reference Path} The global reference path from the starting position to the goal is generated using the A* algorithm, same as \cite{thirugnanam2022safety}. 
To track the path, an initial reference speed of $0.2$ is assigned. 
The local reference path $\mathbf{x}_{r,t+k}, k=0,\dots,N$ is then computed as the product of the reference speed, the prediction horizon $N$, and the sampling time $\Delta t$. 
After each update, the reference speed is reset to the current robot speed.
\subsubsection{DHOCBF}
The candidate DHOCBF function $\tilde{\psi}_{0}(\mathbf{x}_{t}|\bar{\mathbf x}_{t})$ is defined by Eq. \eqref{eq:DCBFs}. Based on Sec. \ref{subsec: Complexity Analysis}, we set $m_{\text{cbf}}=1 $ and hyperparameter $\gamma_{1}=0.1$.
\subsubsection{MPC Design}
The cost function of the MPC problem \eqref{eq:impc-dcbf} consists of current cost
$\sum_{k=0}^{N-1} q(\mathbf{x}_{t,k}^j,\mathbf{u}_{t,k}^{j},\omega_{t,k,i}^{j,i_{o}})= \sum_{k=0}^{N-1} (||\mathbf{x}_{t,k}^{j}-\mathbf{x}_{r,t+k}||_Q^2 + ||\mathbf{u}_{t,k}^{j}-\mathbf{u}_{r}||_R^2 +||\omega_{t,k,i}^{j,i_{o}}-\omega_{r}||_S^2)$, and terminal cost $p(\mathbf{x}_{t,N}^{j})=||\mathbf{x}_{t,N}^{j}-\mathbf{x}_{r,t+N}||_P^2.$ $N=12.$
\subsubsection{Convergence Criteria}
We use the following absolute and relative convergence functions as convergence criteria mentioned in Fig. \ref{fig:iteration-module}:
\begin{small}
\begin{equation}
\label{eq:convergence-criteria}
    \begin{split}
    e_{\text{abs}}(\mathbf{X}_{t}^{*, j}, \mathbf{U}_{t}^{*, j}) &= ||\mathbf{X}^{*, j} - \bar{\mathbf{X}}^{*, j}|| \\
    e_{\text{rel}}(\mathbf{X}_{t}^{*, j}, \mathbf{U}_{t}^{*, j}, \bar{\mathbf{X}}_{t}^{j}, \bar{\mathbf{U}}_{t}^{j}) &= ||\mathbf{X}^{*, j} - \bar{\mathbf{X}}^{*, j}||/||\bar{\mathbf{X}}^{*, j}||.
\end{split}
\end{equation}
\end{small}
The iterative optimization stops when $e_{\text{abs}} < \varepsilon_{\text{abs}}$ or $e_{\text{rel}} < \varepsilon_{\text{rel}}$, where $\varepsilon_{\text{abs}} = 0.05$, $\varepsilon_{\text{rel}} = 10^{-2}$ and the maximum iteration number is set as $j_{\text{max}} = 50$.

\begin{figure*}[!t]
    \vspace*{0.2cm}
    \centering
    \begin{subfigure}[t]{0.23\linewidth}
        \centering
    \includegraphics[width=1.0\linewidth]{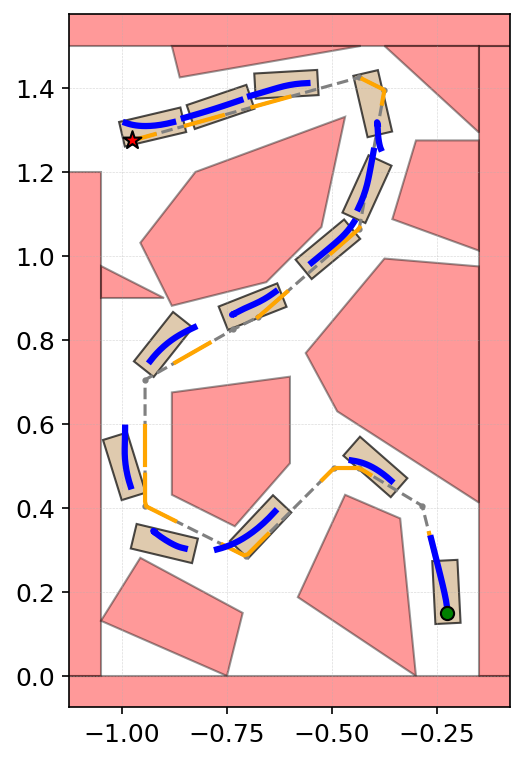}
        \caption{One rectangle robot}
        \label{subfig:1}
    \end{subfigure}
    \begin{subfigure}[t]{0.23\linewidth}
        \centering
        \includegraphics[width=1.0\linewidth]{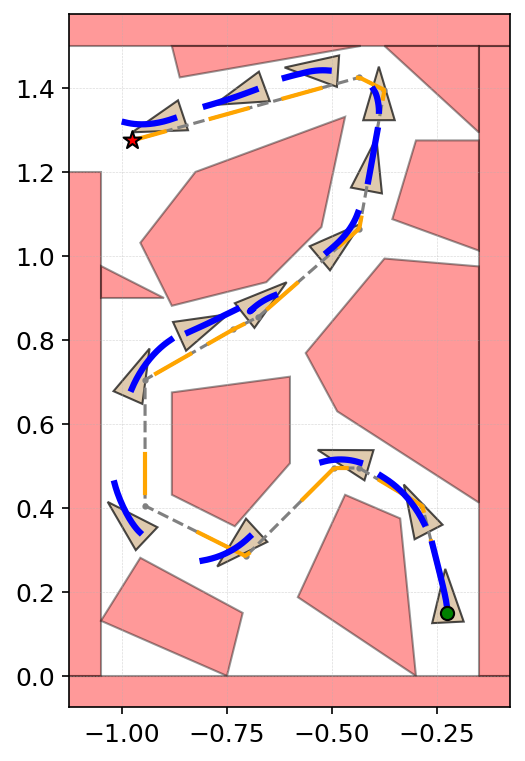}
        \caption{One triangle robot}
        \label{subfig:2}
    \end{subfigure}  
    \begin{subfigure}[t]{0.23\linewidth}
        \centering
        \includegraphics[width=1.0\linewidth]{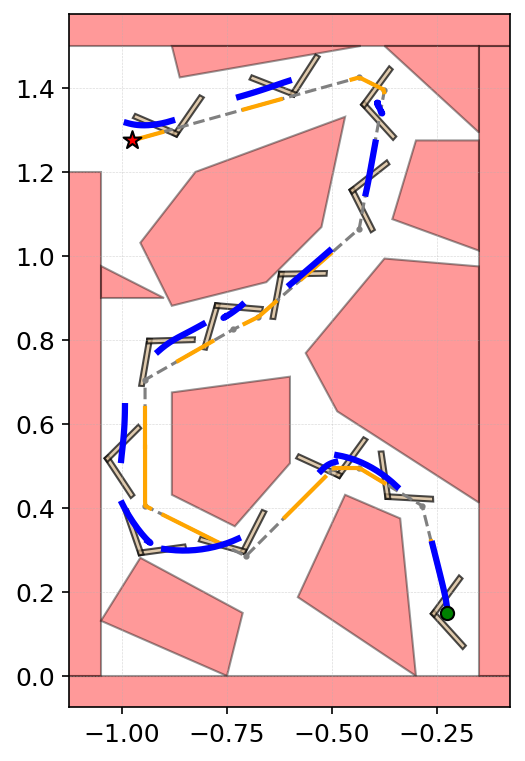}
        \caption{One L-shaped (nonconvex) robot}
        \label{subfig:3}
    \end{subfigure}
        \begin{subfigure}[t]{0.224\linewidth}
        \centering
        \includegraphics[width=1.0\linewidth]{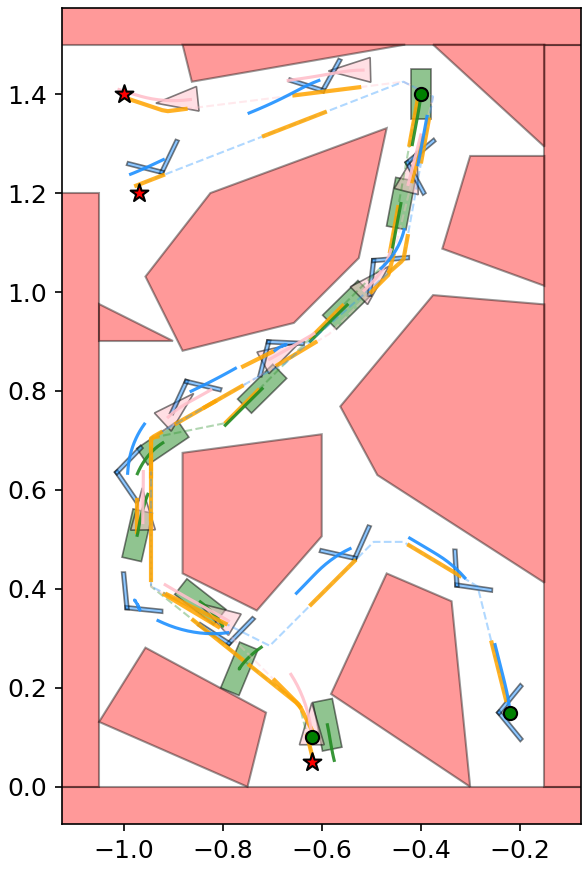}
        \caption{Team of 3 robots}
        \label{subfig:4}
    \end{subfigure}
    \caption{Autonomous navigation results for different robot geometries in 2-D environments. The green circle and the red pentagon denote the starting position and the goal positions, respectively.  
The orange curve represents the local reference path, while the curves in other colors correspond to the predicted local trajectories generated by the controller.}
    \label{fig: 2-D}
\end{figure*}

\subsubsection{Safe Trajectory Generation}
We study autonomous navigation in cluttered maze environments using three different robot geometries. 
From Fig. \ref{fig: 2-D}, we observe tight obstacle-avoidance maneuvers to escape potential deadlocks, demonstrating both the safety and planning capabilities of the proposed method. 

\subsection{Three robots in a 2-D environment}
\label{subsec: 2-D-mlti}

Here we consider the three robots described above (rectangle, triangle, and L-shaped), but scaled down to  two-thirds of their original size. We consider the same environment.  
Fig. \ref{subfig:4} shows that even when three robots encounter each other in narrow spaces, they can safely avoid one another and successfully reach their respective goals. 
In particular, the rectangle robot is observed to temporarily reverse its motion to yield space and avoid a potential conflict with the L-shape robot. 
These results further validate the effectiveness of our approach in handling both static and dynamic obstacles.

\subsection{3-D Numerical Setup}
\label{sec:3D-sim}

\subsubsection{System Dynamics}
Consider a 3-D discrete-time unicycle model \cite{lin20153} in the form
\begin{small}
\begin{equation}
\label{eq:3d-unicycle-model}
\begin{bmatrix}
x_{t+1}-x_t \\
y_{t+1}-y_t \\
z_{t+1}-z_t \\
\theta_{1,t+1}-\theta_{1,t} \\
\theta_{2,t+1}-\theta_{2,t} \\
v_{t+1}-v_t
\end{bmatrix}
=
\begin{bmatrix}
v_t \cos(\theta_{2,t})\cos(\theta_{1,t}) \Delta t \\
v_t \cos(\theta_{2,t})\sin(\theta_{1,t}) \Delta t \\
v_t \sin(\theta_{2,t}) \Delta t \\
0 \\
0 \\
0
\end{bmatrix}
+
\begin{bmatrix}
0 \\
0 \\
0 \\
u_{1,t}\,\Delta t \\
u_{2,t}\,\Delta t \\
u_{3,t}\,\Delta t
\end{bmatrix},
\end{equation}
\end{small}

where $\mathbf{x}_{t}=[x_{t},\,y_{t},\,z_{t},\,\theta_{1,t},\,\theta_{2,t},\,v_{t}]^{\top}$ 
represents the 3-D position, yaw angle, pitch angle, and linear speed, 
and $\mathbf{u}_{t}=[u_{1,t},\,u_{2,t},\,u_{3,t}]^{\top}$ 
denotes yaw rate, pitch rate, and linear acceleration. 
The sampling time is $\Delta t=0.1$. 
System~\eqref{eq:3d-unicycle-model} is subject to the following state and input constraints:
\begin{small}
\begin{equation}
\begin{split}
\label{eq:state-input-constraint2}
\mathcal{X}
&=
\left\{
\mathbf{x}_{t}\in \mathbb{R}^{6}:
-4\cdot \mathcal{I}_{6\times1}
\le
\mathbf{x}_{t}
\le
4\cdot \mathcal{I}_{6\times1}
\right\},\\
\mathcal{U}
&=
\left\{
\mathbf{u}_{t}\in \mathbb{R}^{3}:
-0.5\cdot \mathcal{I}_{3\times1}
\le
\mathbf{u}_{t}
\le
0.5\cdot \mathcal{I}_{3\times1}
\right\}.
\end{split}
\end{equation}
\end{small}
\subsubsection{System Configuration}
The L-shape robot consists of two perpendicular arms forming a right angle, 
with the reference point located at the centroid. 
The planar footprint is extruded along the $z$-axis by $\pm 0.03\,\mathrm{m}$, 
resulting in a total height of $0.06\,\mathrm{m}$.
The initial state is 
$\mathbf{x}_0=[0.225,\,0.338,\,0.135,\,0.0264,\,0.769,\,0.0]^\top$, 
with the initial orientation aligned with the first guiding path segment. 
The target position is 
$\mathbf{x}_T=[2.99925,\,1.4625,\,0.99]^\top$. 
The reference inputs are 
$\mathbf{u}_r=[0,0,0]^\top$ and $\Omega_r=[1,1,1]^\top$.
All other settings are identical to the 2-D case in Sec. \ref{subsec: 2-D}, including reference path, DHOCBF, MPC design, convergence criteria, and solver configuration, except that $N=8.$
\subsubsection{Safe Trajectory Generation} 
\begin{figure}[!t]
    \vspace*{0.2cm}
    \centering
    \begin{subfigure}[t]{0.75\linewidth}
        \centering
    \includegraphics[width=1.0\linewidth]{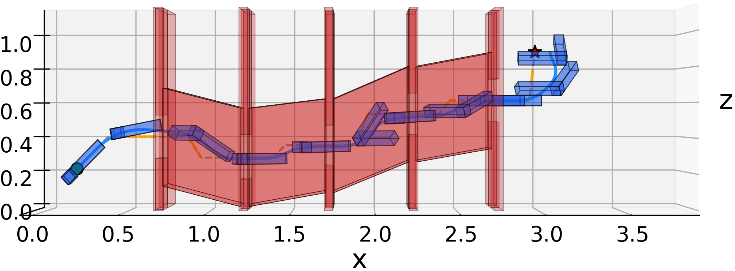}
        \caption{Front view}
        \label{subfig:5}
    \end{subfigure}
    \begin{subfigure}[t]{0.65\linewidth}
        \centering
        \includegraphics[width=1.0\linewidth]{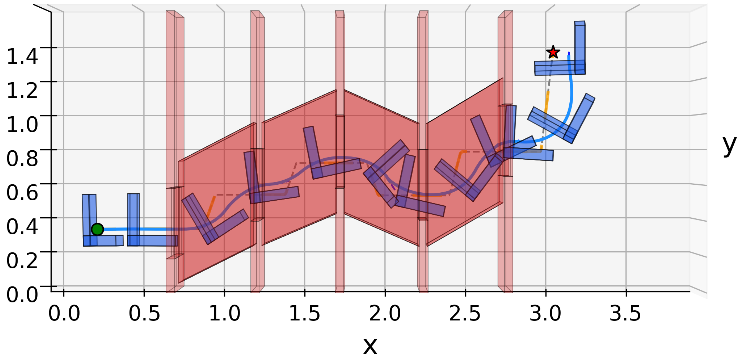}
        \caption{Top View}
        \label{subfig:6}
    \end{subfigure}
        \begin{subfigure}[t]{0.55\linewidth}
        \centering
        \includegraphics[width=1.0\linewidth]{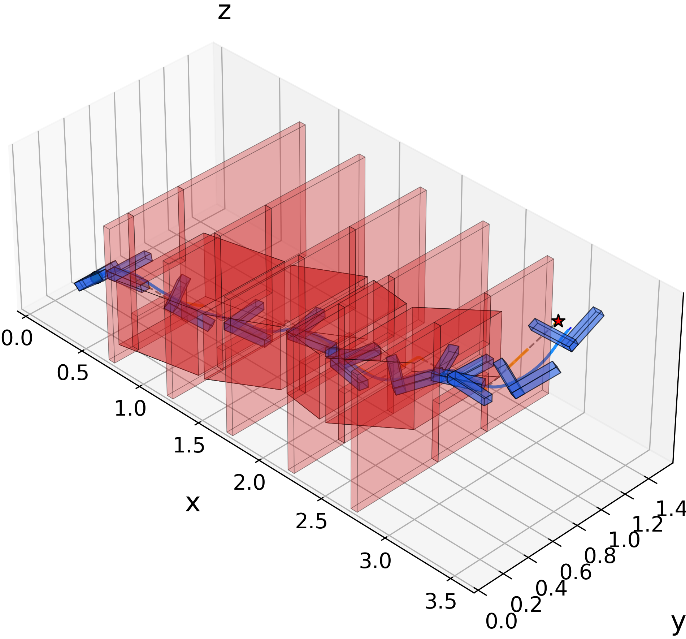}
        \caption{Isometric view}
        \label{subfig:7}
    \end{subfigure}
    \caption{Navigation results from three viewpoints. 
The robot moves from the green circle (start) to the red pentagram (goal), generating collision-free trajectories in the cluttered environment.}
    \label{fig: 3-D}
\end{figure}

As shown in Fig.~\ref{fig: 3-D}, we construct a cluttered 3-D maze environment consisting of five walls. 
Each wall contains a narrow opening slightly larger than the robot, and adjacent walls are connected by narrow passages that are also slightly wider than the openings. 
The L-shaped robot navigates through the 3-D environment by coordinated translations and rotations, avoiding collisions and deadlocks while successfully reaching the goal.
\subsubsection{Computational Efficiency}

\begin{table}[]
\centering
\vspace*{3mm}
\resizebox{0.45\textwidth}{!}{
\begin{tabular}{|c|c|c|cc|}
\hline
\multicolumn{3}{|c|}{Obstacle shapes} & $\gamma_{1} = 0.1$ & $\gamma_{1} = 0.2$ \\ \hline

\multirow{3}{*}{2-D Rectangle} 
& N=6  & mean / std (ms) & $5.016\pm0.760$ & $4.819\pm0.675$  \\ \cline{2-5}

& N=12 & mean / std (ms) & $11.406\pm2.481$ & $10.917\pm2.064$ \\
 \cline{2-5}

& N=24 & mean / std (ms) & $30.592\pm8.881$ & $31.278\pm9.042$ \\ \hline

\multirow{3}{*}{2-D Triangle} 
& N=6  & mean / std (ms) & $4.326\pm0.432$ & $4.457\pm0.659$ \\ \cline{2-5}

& N=12 & mean / std (ms) & $10.254\pm1.184$ & $10.318\pm2.099$ \\ \cline{2-5}

& N=24 & mean / std (ms) & $25.212\pm5.186$ & $25.065\pm7.107$\\ \hline

\multirow{3}{*}{2-D L-shape} 
& N=6  & mean / std (ms) & $9.755\pm1.555$ & $9.822\pm1.452$ \\ \cline{2-5}

& N=12 & mean / std (ms) & $23.326\pm3.829$ & $22.634\pm3.936$ \\ \cline{2-5}

& N=24 & mean / std (ms) & $58.877\pm13.981$ & $65.218\pm10.134$\\ \hline

\multirow{3}{*}{3-D L-shape} 
& N=6  & mean / std (ms) & $6.551\pm1.003$ & $6.732\pm1.592$ \\ \cline{2-5}

& N=12 & mean / std (ms) & $13.786\pm2.981$ & $14.488\pm6.946$ \\ \cline{2-5}

& N=24 & mean / std (ms) & $28.124\pm4.894$ & $28.213\pm4.828$\\ \hline

\end{tabular}
}
\caption{Computation time statistics for 2-D and 3-D robot shapes under different horizon $N$ and DHOCBF hyperparameter $\gamma_1$. Each entry reports mean and standard deviation over 50 trials.}
\label{tab:compuation-time1}
\end{table}
For each robot shape in both the 2-D and 3-D settings, 50 random initial locations are generated within the free space. The system is then simulated for 20 time steps, and the per-step computation time (in milliseconds) is recorded, as summarized in Table \ref{tab:compuation-time1}. As expected, the computation time increases with the prediction horizon, but remains fast in all cases. The 3-D L-shape robot exhibits shorter computation times than its 2-D counterpart due to fewer obstacles within its sensing region. Moreover, the proposed method achieves faster computation compared to \cite{thirugnanam2022safety} under the same map and robot geometries.
\section{Conclusion and Future Work}
\label{sec:Conclusion and Future Work}
In this paper, we presented an iterative convex MPC-DHOCBF framework for obstacle avoidance among polytopic sets. 
By combining closest-point supporting hyperplanes with DHOCBFs, the method preserves convexity at each iteration while ensuring safety. 
The framework accommodates robots modeled as unions of convex polytopes, enabling nonconvex geometries such as L-shaped bodies, and extends naturally to multi-robot systems and three-dimensional environments. 
Computational experiments show consistent millisecond-level solve times, demonstrating real-time capability. 
Future work will investigate robustness under modeling uncertainty and formal safety guarantees under geometry linearization.
\bibliographystyle{IEEEtran}
\balance
\bibliography{references.bib}

@article{xiao2021high,
  title={High-order control barrier functions},
  author={Xiao, Wei and Belta, Calin},
  journal={IEEE Trans. Autom. Control},
  volume={67},
  number={7},
  pages={3655--3662},
  year={2021},
}

@article{ames2016control,
  title={Control barrier function based quadratic programs for safety critical systems},
  author={Ames, Aaron D and Xu, Xiangru and Grizzle, Jessy W and Tabuada, Paulo},
  journal={IEEE Trans. Autom. Control},
  volume={62},
  number={8},
  pages={3861--3876},
  year={2016},
}

@inproceedings{nguyen2016exponential,
  title={Exponential control barrier functions for enforcing high relative-degree safety-critical constraints},
  author={Nguyen, Quan and Sreenath, Koushil},
  booktitle={2016 Amer. Control Conf. (ACC)},
  pages={322--328},
  year={2016},
}

@inproceedings{liu2023auxiliary,
  title={Auxiliary-Variable Adaptive Control Barrier Functions for Safety Critical Systems},
  author={Liu, Shuo and Xiao, Wei and Belta, Calin A},
  booktitle={2023 62nd IEEE Conf. Decis. Control (CDC)},
  year={2023}
}

@inproceedings{liu2023iterative,
  title={Iterative convex optimization for model predictive control with discrete-time high-order control barrier functions},
  author={Liu, Shuo and Zeng, Jun and Sreenath, Koushil and Belta, Calin A},
  booktitle={2023 Amer. Control Conf. (ACC)},
  pages={3368--3375},
  year={2023},
}

@inproceedings{khazoom2022humanoid,
  title={Humanoid self-collision avoidance using whole-body control with control barrier functions},
  author={Khazoom, Charles and Gonzalez-Diaz, Daniel and Ding, Yanran and Kim, Sangbae},
  booktitle={2022 IEEE-RAS 21st Int. Conf. Humanoid Robots (Humanoids)},
  pages={558--565},
  year={2022},
}

@article{xiong2022discrete,
  author={Xiong, Yuhan and Zhai, Di-Hua and Tavakoli, Mahdi and Xia, Yuanqing},
  journal={IEEE Trans. Cybern.}, 
  title={Discrete-Time Control Barrier Function: High-Order Case and Adaptive Case}, 
  year={2022},
  volume={},
  number={},
  pages={1-9},
}

@inproceedings{agrawal2017discrete,
  title={Discrete Control Barrier Functions for Safety-Critical Control of Discrete Systems with Application to Bipedal Robot Navigation.},
  author={Agrawal, Ayush and Sreenath, Koushil},
  booktitle={Robotics: Science and Systems},
  volume={13},
  year={2017},
  organization={Cambridge, MA, USA}
}

@article{stellato2020osqp,
  title={{OSQP}: An operator splitting solver for quadratic programs},
  author={Stellato, Bartolomeo and Banjac, Goran and Goulart, Paul and Bemporad, Alberto and Boyd, Stephen},
  journal={Math. Program. Comput.},
  volume={12},
  number={4},
  pages={637--672},
  year={2020},
  publisher={Springer}
}

@inproceedings{peng2023safe,
  title={Safe bipedal path planning via control barrier functions for polynomial shape obstacles estimated using logistic regression},
  author={Peng, Chengyang and Donca, Octavian and Castillo, Guillermo and Hereid, Ayonga},
  booktitle={2023 IEEE Int. Conf. Robot. Autom. (ICRA)},
  pages={3649--3655},
  year={2023}
}

@inproceedings{thirugnanam2022duality,
  title={Duality-based convex optimization for real-time obstacle avoidance between polytopes with control barrier functions},
  author={Thirugnanam, Akshay and Zeng, Jun and Sreenath, Koushil},
  booktitle={2022 Amer. Control Conf. (ACC)},
  pages={2239--2246},
  year={2022},
}

@article{wu2025optimization,
  title={Optimization-Free Smooth Control Barrier Function for Polygonal Collision Avoidance},
  author={Wu, Shizhen and Fang, Yongchun and Sun, Ning and Lu, Biao and Liang, Xiao and Zhao, Yiming},
  journal={IEEE Trans. Cybern.},
  year={2025},
}

@inproceedings{thirugnanam2022safety,
  title={Safety-critical control and planning for obstacle avoidance between polytopes with control barrier functions},
  author={Thirugnanam, Akshay and Zeng, Jun and Sreenath, Koushil},
  booktitle={2022 IEEE Int. Conf. Robot. Autom. (ICRA)},
  pages={286--292},
  year={2022},
}

@inproceedings{liu2025safety,
  title={Safety-critical planning and control for dynamic obstacle avoidance using control barrier functions},
  author={Liu, Shuo and Mao, Yihui and Belta, Calin A},
  booktitle={2025 Amer. Control Conf. (ACC)},
  pages={348--354},
  year={2025},
}

@inproceedings{ames2019control,
  title={Control barrier functions: Theory and applications},
  author={Ames, Aaron D and Coogan, Samuel and Egerstedt, Magnus and Notomista, Gennaro and Sreenath, Koushil and Tabuada, Paulo},
  booktitle={2019 18th Eur. Control Conf. (ECC)},
  pages={3420--3431},
  year={2019},
}

@article{liu2025auxiliary,
  title={Auxiliary-variable adaptive control barrier functions},
  author={Liu, Shuo and Xiao, Wei and Belta, Calin A},
  journal={arXiv preprint arXiv:2502.15026},
  year={2025}
}

@article{liu2025learning,
  title={Learning-Enabled Iterative Convex Optimization for Safety-Critical Model Predictive Control},
  author={Liu, Shuo and Huang, Zhe and Zeng, Jun and Sreenath, Koushil and Belta, Calin A},
  journal={IEEE Open J. Control Syst.},
  year={2025},
}

@article{chen2017obstacle,
  title={Obstacle avoidance for low-speed autonomous vehicles with barrier function},
  author={Chen, Yuxiao and Peng, Huei and Grizzle, Jessy},
  journal={IEEE Trans. Control Syst. Technol.},
  volume={26},
  number={1},
  pages={194--206},
  year={2017},
}

@article{verginis2019closed,
  title={Closed-form barrier functions for multi-agent ellipsoidal systems with uncertain lagrangian dynamics},
  author={Verginis, Christos K and Dimarogonas, Dimos V},
  journal={IEEE Control Syst. Lett.},
  volume={3},
  number={3},
  pages={727--732},
  year={2019},
}

@article{singletary2022safety,
  title={Safety-critical manipulation for collision-free food preparation},
  author={Singletary, Andrew and Guffey, William and Molnar, Tamas G and Sinnet, Ryan and Ames, Aaron D},
  journal={IEEE Robot. Autom. Lett.},
  volume={7},
  number={4},
  pages={10954--10961},
  year={2022},
}

@article{wei2024diffocclusion,
  title={Diffocclusion: Differentiable optimization based control barrier functions for occlusion-free visual servoing},
  author={Wei, Shiqing and Dai, Bolun and Khorrambakht, Rooholla and Krishnamurthy, Prashanth and Khorrami, Farshad},
  journal={IEEE Robot. Autom. Lett.},
  volume={9},
  number={4},
  pages={3235--3242},
  year={2024},
}

@inproceedings{chen2025control,
  title={Control barrier functions via minkowski operations for safe navigation among polytopic sets},
  author={Chen, Yi-Hsuan and Liu, Shuo and Xiao, Wei and Belta, Calin and Otte, Michael},
  booktitle={2025 IEEE 64th IEEE Conf. Decis. Control (CDC)},
  pages={4481--4488},
  year={2025},
}

@article{lin20153,
  title={{3-D} velocity regulation for nonholonomic source seeking without position measurement},
  author={Lin, Jinbiao and Song, Shiji and You, Keyou and Wu, Cheng},
  journal={IEEE Trans. Control Syst. Technol.},
  volume={24},
  number={2},
  pages={711--718},
  year={2015},
}

\end{document}